\documentclass{article}

\usepackage[square,numbers]{natbib}
\usepackage{amsmath}
\usepackage{graphicx}
\usepackage{caption}
\usepackage{subcaption}

\DeclareMathAlphabet\mathbfcal{OMS}{cmsy}{b}{n}

\usepackage[utf8]{inputenc} 
\usepackage[T1]{fontenc}    
\usepackage{hyperref}       
\usepackage{url}            
\usepackage{booktabs}       
\usepackage{amsfonts}       
\usepackage{nicefrac}       
\usepackage{microtype}      
\usepackage{authblk}

\author[1]{Huangjie Zheng}
\author[1]{Jiangchao Yao}
\author[1]{Ya Zhang}
\author[2]{Ivor W. Tsang}
\author[1]{Jia Wang}

\affil[1]{ Cooperative Medianet Innovation Center \\ Shanghai Jiao Tong University}
\affil[2]{Centre for Artificial Intelligence \\ University of Technology Sydney}
\affil[1]{ \texttt{\{zhj865265, sunarker, ya\_zhang, jiawang\}@sjtu.edu.cn}}
\affil[2]{\texttt{ivor.tsang@uts.edu.au}}
\date{}

\title{Understanding VAEs in Fisher-Shannon Plane}

\begin{document}

\maketitle
 
\begin{abstract}
In information theory, Fisher information and Shannon information (entropy) are respectively used to quantify the uncertainty associated with the distribution modeling and the uncertainty in specifying the outcome of given variables. These two quantities are complementary and are jointly applied to information behavior analysis in most cases. The uncertainty property in information asserts a fundamental trade-off between Fisher information and Shannon information, which enlightens us the relationship between the encoder and the decoder in \emph{variational auto-encoders} (VAEs). In this paper, we investigate VAEs in the Fisher-Shannon plane, and demonstrate that the representation learning and the log-likelihood estimation are intrinsically related to these two information quantities. Through extensive qualitative and quantitative experiments, we provide with a better comprehension of VAEs in tasks such as high-resolution reconstruction, and representation learning in the perspective of Fisher information and Shannon information. We further propose a variant of VAEs, termed as Fisher auto-encoder (FAE), for practical needs to balance Fisher information and Shannon information. Our experimental results have demonstrated its promise in improving the reconstruction accuracy and avoiding the non-informative latent code as occurred in previous works. 
\end{abstract}

\section{Introduction}
The common latent variable models fit $p_\theta(x,z) = p_\theta(z)p_\theta(x|z)$  in order to model the data $x$ with a latent variable $z$ as representation. Variational Autoencoders (VAEs) \cite{journals/corr/KingmaW13}, recently as one of the most popular latent variable models, maximize the evidence lower bound (ELBO) in an encoding/decoding mechanism. 
$$
\mathcal{L} = \underset{x \sim data}{\mathbf{E}}\left[ \underset{z \sim q_\phi}{\mathbf{E}} \left[ \log p_\theta(x|z) \right ] - \mathcal{D}_{\text{KL}}\left( q_\phi(z|x)|| p(z) \right) \right]
$$
where $p_\theta(x|z)$ and $q_\phi(z|x)$ are encoder and decoder implemented with neural networks parameterized by $\theta, \phi$; $\mathcal{D}_{\text{KL}}$ denotes the Kullback-Leibler divergence. The learning targets of VAEs may be interpreted in the following two perspectives. In the perspective of variational optimization \cite{optimizationVAE}, VAEs aim to learn a proper model by maximum likelihood with $p_\theta(x|z)$.  
In the perspective of representation learning~\cite{representationlearning}, VAEs target to learn latent codes that are sufficiently encoded with information about the input by $q_\phi(z|x)$. 
Several variants of VAEs~\cite{DBLP:journals/corr/OordKK16,DBLP:journals/corr/OordKVEGK16,chen2016variational,zhao2017infovae} have been developed based on above two perspectives in recent years.

In recent studies, it has been reported that VAEs are difficult to balance between the representation learning and likelihood maximizing. For instance, the latent variable generated by the encoder is approximately ignored when the decoder is too expressive \cite{bowman2015generating}.  As discussed in \cite{alemifixing}, the evidence lower bound (ELBO) lacks a measure in the quality of the representation, since the KL divergence $\mathcal{D}_{\text{KL}}\left( q_\phi(z|x)|| p(z) \right)$ only controls the way VAE encodes a representation.  Several studies in VAEs~\cite{chen2016variational,zhao2017infovae} have attempted to improve the balance between the learned representation quality and ELBO maximization on the basis of information theory. A Shannon entropy-based constraint is introduced to assess the quality of representation learning when optimizing the ELBO, so as to guarantee that sufficient information of the observation flows into the latent code. For example, \cite{phuong2018the} optimized ELBO plus a mutual information regularizer to explicitly control the information stored in latent codes. Although it is useful to consider mutual information (a member of Shannon family) in VAE encoding for representation learning, how these information quantities affect VAEs has not yet been theoretically analyzed so far. Besides, previous works mainly leverage the Shannon information, which usually suffers from the intractability in computing, yielding an approximation surrogate~\cite{phuong2018the}.  

In information theory, the uncertainty property~\cite{104312,VIGNAT200327} has revealed a trade-off between Fisher information and Shannon information, which quantify the uncertainty associated with distribution modeling and the entropy in the predicted values of variables respectively~\cite{rosso2015noise}. Fisher-Shannon (FS) information plane~\cite{VIGNAT200327} is proposed to analyze the complementarity between Fisher information and Shannon information. 
In this paper, based on the uncertainty property, we attempt to investigate VAEs in FS information plane. We first perform a theoretical analysis with VAEs and show that representation learning with latent variable models via Maximum likelihood estimation is intrinsically related to the trade-off between Fisher information and Shannon entropy. Based on the above findings, we propose a family of VAEs regularized by Fisher information, named \emph{Fisher auto-encoder} (FAE), to control the information quality during encoding/decoding. 
Finally, we perform a range of experiments to analyze a variety of VAE models in the FS information plane to empirically validate our findings. In addition, regularized with the Fisher information, FAE is shown to not only provide a novel insight in the information trade-off, but can also improve the reconstruction accuracy and the representation learning.

\section{Related work}

\textbf{Information uncertainty:} Fisher information and Shannon information are considered important tools to describe
the informational behavior in information systems respectively in the distribution modeling view and in the variable view \cite{brunel1998mutual}. The generalization of Heisenberg Uncertainty Principle \cite{busch2007heisenberg} into information system demonstrates that Fisher information and Shannon information are intrinsically linked, with the uncertainty property, where higher Fisher information will result in lower Shannon information and vice versa \cite{104312}. With this property, Fisher information and Shannon information are considered complementary aspects and be widely used in solving dual problem when one aspect is intractable \cite{martin1999fisher}. To better take advantage of this property, \cite{VIGNAT200327} construct the Fisher-Shannon information plane for signal analysis in joint view of Fisher-Shannon information. 

\textbf{VAEs in information perspective:} Variational autoencoders \cite{journals/corr/KingmaW13}, with a auto-encoding form, can be regarded as an information system serves two goals. On one hand, we expect a proper distribution estimation to maximize the marginal likelihood; on the other hand, we hope the latent code can provide sufficient information of data point so as to serve downstream tasks. To improve the log-likelihood estimation, several works, such as PixelCNN and PixelRNN \cite{DBLP:journals/corr/OordKK16,DBLP:journals/corr/OordKVEGK16} model the dependencies among pixels with autoregressive structure to achieve an expressive density estimator. As for the latent code, plenty of works address it in the perspective of Shannon information \cite{chen2016variational}. Mutual information, an member of Shannon family, is applied to measure the mutual dependence between datapoint $x$ and latent code $z$ \cite{zhao2017infovae,alemifixing}. The leverage of mutual information is achieved with Maximum-Mean Discrepancy \cite{zhao2017infovae} and Wasserstein distance \cite{tolstikhin2018wasserstein}. More generally, \cite{phuong2018the} regularize the mutual information in VAE's objective to control the information in latent code.   

\textbf{Effects of Fisher information and Shannon information:} Fisher information and Shannon information (typically we call entropy), as complementary aspects, possess their properties. In \cite{rosso2015noise}, the entropy is explained as a measure of ``global character” that is invariant to strong changes in the distribution, while the Fisher information is interpreted as a measure of the ability to model a distribution, which corresponds to the ``local" characteristics. The characteristics of these two sides have been taken into advantages of several existing works, \textit{e.g.}, entropy has been introduced in improving deep neural networks on tasks like classification \cite{silva2005neural,michael2018on}; FI has been introduced to to evaluate neural network's parameter estimation \cite{7472157,Desjardins:2015:NNN:2969442.2969471}. To better understand how these two perspectives affect the mechanism of VAEs, we study VAEs in the Fisher-Shannon information plane to provide a complete understanding of VAEs in a joint view of Fisher information and Shannon information.

\section{Fisher-Shannon Information Plane}
In this section, we first present the information uncertainty property to link the Fisher information and the Shannon information~\cite{104312} of the random variable. After that, a Fisher-Shannon information plane~\cite{VIGNAT200327} is constructed to jointly analyze the characteristics of the random variable on its distribution. This provides the simple basics to understand VAEs with the Fisher-Shannon information plane.

\subsection{Fisher-Shannon Information Uncertainty property}\label{sec:ineq}
In information theory, considering a random variable $X$, whose probability density function is denoted as $f(x)$, the Fisher information\footnote{Note that, we follow the non-parametric Fisher information definition that differentiates on random variables, which can be  transformed with a translation of parameter from parametric version \cite{STAM1959101}.} for $X$ and its Shannon entropy can be formulated as:
\begin{align} \label{eq:fiShannon}
\begin{split}
&\text{Fisher Information:} \quad \mathcal{J}(X) = \int_x \left( \frac{\partial}{\partial x} {f(x)} \right)^2 \frac{dx}{f(x)} \\
&\text{Shannon Entropy:} \quad \mathcal{H}(X) = - \int_x f(x)\log f(x) dx
\end{split}
\end{align}
Above two information quantities are respectively related to the precision that the model fits in observed data and the uncertainty in the outcome of the random variable. For convenience to use in deduction, Shannon entropy is usually transformed to the following quantity which is called the entropy power~\cite{STAM1959101}:
\begin{align} \label{eq:Shannon}
\begin{split}
&\mathcal{N}(X) = \frac{\exp( 2 \mathcal{H}(X))}{2\pi \exp(1)}
\end{split}
\end{align}
The measure of $\mathcal{N}(X)$ and $\mathcal{J}(X)$ verifies a set of resembling inequalities in information theory \cite{STAM1959101}. Specifically, one of the inequalities connecting the two quantities and being tightly related to the phenomena studied in VAEs, is the uncertainty inequality~\cite{104312}, which is formulated as:
\begin{align} \label{eq:UI}
\begin{split}
&\mathcal{N}(X) \mathcal{J}(X) \geq 1
\end{split}
\end{align}
where the equality holds when the random variable $X$ is a Gaussian variable. Note that this inequality possesses several versions; in the case of a random vector $\mathbfcal{X}=(X_1,X_2,...,X_n)$, the corresponding Fisher information turns into a $n \times n$ dimensional Fisher information matrix and we need to compute the trace of this matrix $tr(\mathbfcal{J}(\mathbfcal{X}))$ (see \cite{1990demboineq}) and we have $\mathcal{N}(\mathbfcal{X}) \cdot tr(\mathbfcal{J}(\mathbfcal{X})) \geq n$. 

 When the distribution of given variable is fixed, the product of the Fisher information $tr(\mathbfcal{J}(\mathbfcal{X}))$ and the entropy power $\mathcal{N}(\mathbfcal{X})$ is a constant that is greater or equal to $1$, which depends on the distribution form, the dimension of the random vector, \textit{etc.} We can further formulate this property as follow:
\begin{align} \label{eq:generalized UI}
\begin{split}
&\mathcal{N}(\mathbfcal{X}) \cdot tr(\mathbfcal{J}(\mathbfcal{X})) = K
\end{split}
\end{align}
where $K$ is a constant number and $K \geq 1$. Eq.~\eqref{eq:UI} and~\eqref{eq:generalized UI} indicate the measure of Fisher information and Shannon information exists a trade-off between these two quantities.

\subsection{Fisher-Shannon Information Plane}\label{sec:plane}
To facilitate the analysis of above two information quantities together, an information plane based on the Fisher information and the Shannon entropy power is proposed in \cite{VIGNAT200327} and we generalize it as follows,
\begin{align}\label{eq:fsplane}
\begin{split}
\mathcal{D}= \{ & \left( \mathcal{N}(\mathbfcal{X}), tr(\mathbfcal {J}(\mathbfcal{X})) \right)| \  \mathcal{N}(\mathbfcal{X}) \geq 0 , tr(\mathbfcal{J}(\mathbfcal{X}))  \geq 0 \text{ and } \\
& \mathcal{N}(\mathbfcal{X}) \cdot tr(\mathbfcal{J}(\mathbfcal{X}))  \geq 1  \}.
\end{split}
\end{align} 
where $\mathcal{D}$ denotes a region $\subset \mathbb{R}^2$, which is limited by the Gaussian case. This plane consists of several Fisher-Shannon (FS) curves $\mathcal{N}(\mathbfcal{X}) \cdot tr(\mathbfcal{J}(\mathbfcal{X})) = K$, which characterizes the random variable with different distributions. 

As discussed in \cite{alemifixing}, the quality of latent variable is hard to measure in maximizing ELBO, and various VAEs, like \cite{chen2016variational,higgins2016beta} have been proposed to balance the trade-off between representation learning and optimization. In the FS plane, different VAEs can be analyzed jointly with Fisher information and Shannon information. By observing their location in FS plane, we can identify the characteristic of this VAE model.

In addition, from the uncertainty property between Fisher information and Shannon information, these two quantities are shown tightly connected. As shown in Eq. \eqref{eq:generalized UI}, when the distribution of given random variable is fixed, the Fisher information and Shannon entropy power's product is a constant, where the trade-off exists. We can take advantage of this trade-off to avoid the intractability in Shannon information computing. In this paper, we propose a family of VAEs that control the Fisher information, named Fisher Auto-Encoder (FAE), which allows a more accurate description in situations where the Shannon information shows limited dynamics \cite{martin1999fisher} in VAEs. The details of FAE will be discussed in the next section.

\section{The Fisher Auto-Encoder}
As shown in the previous section, one can apply either Fisher information or Shannon entropy power to control the trade-off between the likelihood estimation $p(x)$ and the dependence between data $x$ and the latent code $z$. In this section, we come up with a family of VAEs that takes advantage of the Fisher information, named Fisher Autoencoder (FAE), and analyze its characteristics in this section. 

\subsection{Fisher Information Control in VAE}
The Fisher AutoEncoder aims to to control the Fisher information quantity in the objective. Thus, the objective becomes to maximize the evidence lower bound (ELBO) with constraint of Fisher information and we reformulate the VAE's objective as follows: 
\begin{align} \label{eq:fielbo}
\begin{split}
 &\underset{{\theta,\phi}}{\text{max}} \quad \underset{x \sim p}{\mathbf{E}}\left[ \underset{z \sim q_\phi}{\mathbf{E}} \left[ \log p(x|z,\theta) \right ] - \mathcal{D}_{\text{KL}}\left( q(z|x,\phi)|| p(z) \right) \right] \\
 &\text{s.t.} \qquad tr(\mathbfcal{J})(x) = F_x,\quad tr(\mathbfcal{J})(z) = F_z 
\end{split}
\end{align}
where $\mathcal{D}_{\text{KL}}$ denotes the Kullback-Leibler divergence \cite{joyce2011kullback}; $F_x$ and $F_z$ are positive constant that denote the desired Fisher information value. A large value of $F_x$ (\textit{resp.} $F_z$) implies we favor a precise distribution estimation in the model parameterized by $\theta$ (\textit{resp.} $\phi$); while a low value of $F_x$ (\textit{resp.} $F_z$) indicates we weaken the distribution modeling to increase the Shannon entropy power.

To solve the scenario described in Eq. \eqref{eq:fielbo}, we transfer this optimization problem into a Lagrangian objective, formulated as:
\begin{align} \label{eq:fae}
\begin{split}
 \mathcal{L}_F(\theta,\phi) & = \underbrace{\underset{x \sim p}{\mathbf{E}}\left[ \underset{z \sim q_\phi}{\mathbf{E}} \left[ \log p(x|z,\theta) \right ] - \mathcal{D}_{\text{KL}}\left( q(z|x,\phi)|| p(z) \right) \right] }_{\text{ELBO}}  \\
 & - \underbrace{\lambda_z \bigg| tr(\mathbfcal{J})(z)  -  F_z \bigg| }_{\text{FI control in encoder}} - \underbrace{ \lambda_x \bigg| tr(\mathbfcal{J})(x)  -  F_x \bigg| }_{\text{FI control in decoder}}
\end{split}
\end{align}
Now the objective consists of three parts, the ELBO to maximize, and two Fisher information regularizers in encoder and decoder, where $\lambda_z$ and $\lambda_x$ are positive constant that control the regularizers. With this objective, we can control the Fisher information in encoder/decoder with an expected desired value $F_{z/x}$. In the most cases, the calculation of Fisher information is not difficult. We can estimate the Fisher information directly by its definition.

\subsection{Characteristic of FAE: an example of FI regularization in Gaussian encoder}\label{sec:var}
Here we give a FAE exemplar that only controls the Fisher information in encoder, which means we set $\lambda_x$ in Eq. \eqref{eq:fae} as zero. In this model, we assume that all random variables are of dimension $1$ (\textit{i.e.} in the scalar case) for simplicity in presentation. The FAE objective is formulated as: 
\begin{align} \label{eq:finalobj}
\begin{split}
 \mathcal{L} &= \underset{x \sim p}{\mathbf{E}}\left[ \underset{z \sim q_\phi}{\mathbf{E}} \left[ \log p(x|z,\theta) \right ] \right] - \underbrace{ \underset{x \sim p}{\mathbf{E}} \left[ \mathcal{D}_{\text{KL}}\left( q(z|x,\phi)|| p(z) \right) 
  \right] -  \lambda_z \bigg| \mathcal{J}(z)  -  F_z \bigg| }_{\mathcal{R}(\phi,\theta)}
\end{split}
\end{align}
This objective consists of a reconstruction term and a generalized regularizor $\mathcal{R}(\phi,\theta)$ that considering the Fisher information other than KL divergence. Same to the VAE \cite{journals/corr/KingmaW13}, both prior distribution $p_\theta(z) = \mathcal{N}(0,1)$ and posterior approximation $q_\phi(z|x)$ are Gaussian, thus the KL-divergence can be analytically computed as:
\begin{align} \label{eq:gausskl}
\begin{split}
- \mathcal{D}_{\text{KL}}\left( q(z|x,\phi)|| p(z) \right) = \frac{1}{2} \left( 1 + \log((\sigma)^2) \right) - (\mu)^2 - (\sigma)^2 
\end{split}
\end{align}
where $\mu$ and $\sigma$ respectively correspond to the mean and standard derivation of a Gaussian distribution. The Fisher information can be easily computed by definition:
\begin{align} \label{eq:FIM}
\begin{split}
 \mathcal{J}(z|x) = \int_z \left( \frac{\partial}{\partial z} {q(z|x)} \right)^2 \frac{dz}{q(z|x)}  =  \frac{1}{ \sigma^2(x) } = \frac{1}{ \sigma^2 }.
\end{split}
\end{align}
Finally, putting Eq.~\eqref{eq:gausskl} and Eq.~\eqref{eq:FIM} together, we have the following regularizer $\mathcal{R}(\phi,\theta)$,
\begin{align} \label{eq:genkl}
\begin{split}
  \mathcal{R}(\phi,\theta) & = -\mathcal{D}_{\text{KL}}\left( q(z|x,\phi)|| p(z) \right)  - \lambda_z\bigg| \mathcal{J}(z)  -  F_z \bigg| \\
   & =  \frac{1}{2} \left( \left( 1 + \log((\sigma)^2) \right) - (\mu)^2 - (\sigma)^2 \right) - \lambda_z  \bigg|\frac{1}{\sigma^2} -  F_z \bigg|
\end{split}
\end{align}
Considering the KL-divergence term in the original VAEs~\cite{journals/corr/KingmaW13}, the optimal is reached at $\sigma^2 = 1$, which aligns the posterior $q_\phi(z|x)$ to a normal distribution $\mathcal{N}(0,1)$. However, in Eq.~\eqref{eq:genkl}, we can observe that the variance is also penalized by the desired Fisher information value $F_z$, which will push the variance to approach zero when $F_z$ is large or make the variance larger than $1$ when $F_z$ is picked as a small value. 

In the above discussion, we analyze the characteristics of FAE in variance control. This property corresponds to the inequality of Cramer-Rao, from which the uncertainty principle shown in Eq. \eqref{eq:UI} can be derive \cite{104312}. Given a stochastic variable $X$ of mean $\mu$ and variance $\sigma^2$, the Fisher information is the lower bound of the variance in a non-biased estimation:
\begin{align} \label{eq:cr}
\begin{split}
 \sigma^2_X \geq \frac{1}{\mathcal{J}(X)},
\end{split}
\end{align}
the equality holds if and only if $X$ is Gaussian. This inequality gives us the first impression of the characteristic of Fisher information: 
When FI is in a low value, the variance of the estimation is forced to be high, causing larger uncertainty of the model estimation. Thus, we need to enlarge the FI to make the variance more controllable.

\subsection{Connection to the Mutual Auto-Encoder}
In this section, we demonstrate the connection between the Fisher Auto-Encoder and  the Mutual Auto-Encoder (MAE) \cite{phuong2018the}, which is representative in the family of VAEs that leverage the Shannon information.

As discussed in the previous section, the product of Fisher information and entropy power is a constant when the distribution of variable is fixed, as shown in Eq. \eqref{eq:generalized UI}. We can derive:

 \begin{align}\label{eq:connection}
 \begin{split}
 log(\mathcal{N}(Z|X)) &= log(K) - log (\mathbfcal{J}(Z|X)) \\
   \Longleftrightarrow log(\mathbfcal{J}(Z|X)) &= - 2\mathcal{H}(Z|X) + constant.
 \end{split}
 \end{align}
 where the FI regularizor in FAE is equivalent to a regularizor of the conditional entropy $\mathcal{H}(Z|X)$.

Looking back into the MAE proposed in \cite{phuong2018the}, this model controls the mutual information between latent variable $z$ and data $x$ as follows: 
\begin{align}\nonumber
\begin{split}
 \mathcal{L} &= \underbrace{ \underset{x \sim p}{\mathbf{E}}\left[ \underset{z \sim q_\phi}{\mathbf{E}} \left[ \log p(x|z,\theta) \right ] - \mathcal{D}_{\text{KL}}\left( q(z|x,\phi)|| p(z) \right) \right]}_{\textbf{ELBO}} \\
 & - \underbrace{C\bigg|\mathcal{I}(x,z)-M\bigg|}_{\text{MI regularizor}}
\end{split}
\end{align}
where $C$ and $M$ are positive constants that respectively control the information regularization and the desired mutual information quantity. Since the mutual information is difficult to compute directly, the mutual information $\mathcal{I}(x,z)$  is inferred using Gibbs inequality \cite{barber2003algorithm}:
\begin{align}\label{eq:micon}
 \begin{split}
 \mathcal{\widehat{I}}(x,z) & = \mathcal{H}(z) - \mathcal{H}(z|x) \\
 & \geq \mathcal{H}(z) +  \underset{x,z \sim p}{\mathbf{E}}\left[ \log r_\omega(z|x) \right]
 \end{split}
 \end{align} 
where $r_\omega(z|x)$ is a parametric distribution that can be modeled by a network. The objective is to maximizing $\mathbf{E}_{x,z}\left[ \log r_\omega(z|x) \right]$ in Eq. \eqref{eq:micon} with the constraint $M$. Thus, MAE intrinsically controls the mutual information by controlling the conditional entropy:
 \begin{align}\label{eq:migap}
 \begin{split}
 \mathcal{I}(x,z) &= \mathcal{H}(x) - \mathcal{H}(x|z) \\
  & = \mathcal{H}(z) - \mathcal{H}(z|x).
 \end{split}
 \end{align} 
with $M$ of large value, the conditional entropy $ \mathcal{H}(Z|X)$ can minimized more to obtain a larger mutual information, and vice versa. FAE can also set constraint $F$ to control the conditional entropy $\mathcal{H}(Z|X)$ (or $\mathcal{H}(X|Z)$).  As Eq. \eqref{eq:connection} shows, using Fisher information, the FI regularizers are equivalent to the regularizers of the conditional entropy; thus the mutual information between $X$ and $Z$ can also be assessed without derive approximative upper or lower bounds. FAE can thus implicitly control the mutual information $\mathcal{I}(X,Z)$ by setting proper Fisher information constraint $F$. 

\section{Experiments}
In this section, we perform a range of experiments to investigate the Fisher-Shannon impacts in VAEs. Meanwhile, we expose how the Fisher Auto-Encoder can improve VAEs in encoding/decoding with the Fisher information constraint.

\subsection{Experiment Goals and Experimental Settings}\label{sec:ex setting}
As discussed, the entropy power and Fisher Information corresponds to different characteristics. Thus, we aim to explore these characteristics and give corresponding analysis in order to give a better understanding of existing VAE variants. Some specific goals are summarized as:
\begin{itemize}
    \item Explore several variants of VAEs models' characteristics in the FS plane. 
    \item Explore the characteristics of latent code \textit{w.r.t.} the Fisher Information and entropy power.
    \item Discuss the effect of different FI constraint in FAE. 
\end{itemize}

The experiments are conducted on the MNIST dataset~\citep{mnist} and the SVHN dataset~\citep{svhn}. The first dataset consists of ten categories of 28$\times$28 hand-written digits and is binarized as in~\citep{larochelle2011neural}. We follow the original partition to split the data as 50,000/10,000/10,000 for the training, validation and test. For SVHN dataset , it is MNIST-like but consists of 32$\times$32 color images. We apply 73257 digits for training, 26032 digits for testing, and 20000 samples from the additional set~\citep{svhn} for validation. Moreover, we also construct a toy dataset using MNIST data to better illustrate the characteristics of FI and entropy power. The dataset consists of 5800 samples of label ``0" and each 100 samples of other labels in training set; the validation and test set remains same as MNIST. 

In FAE and its baselines, all random variables are supposed to be Gaussian variables. Here we only concern the hyper-parameters $\lambda_z$ to adjust the penalty of Fisher information in encoding. In practice, we observe this value can be effective when set from 0.01 to 10 (depends on dataset). For the architecture of inference network and generative network, we both deploy a 5-layers network. Since the impacts of fully-connected and convolution architecture do not differ much in the experiments, we here present results using the architecture as 5 full-connected layers of dimension 300. The latent code is of dimension 40.

\subsection{Quantitative Results}
\subsubsection{Fisher-Shannon Plane Analysis}
In this part, we conduct a series of experiments on different models to evaluate them in  Fisher-Shannon plane to present different characteristics of using Fisher information and entropy power.

\begin{figure}[htb]
 \centering
 \includegraphics[width=\textwidth]{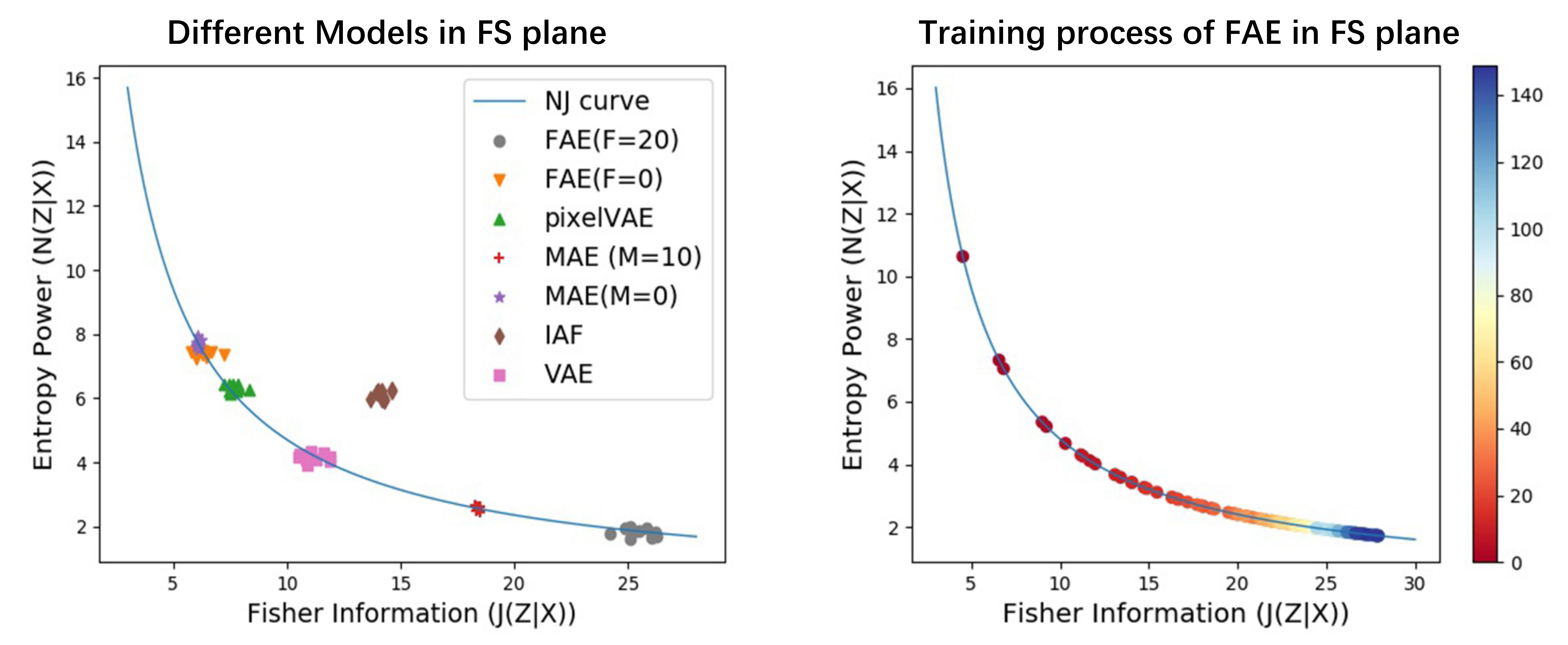}
\caption{(Best view in color) Visualization of variants of VAE models in the Fisher-Shannon Plane (\textbf{left}) and visualization of FAE's training process in the plane. (\textbf{right}) The color bar indicates the training epoch.}\label{fig:FS}
\end{figure}

We first evaluate the test log-likelihood. To compute the test marginal negative log-likelihood (NLL), we apply the Importance Sampling \cite{burda2015importance} with 5,000 data points for the previously mentioned models. We select the most representative average results from extensive hyper-parameter configuration and expose them in Table \ref{tab:nll}: when the Fisher information constraint of $q_{z|x}$ in FAE ($F_z=20$) (or the mutual information constraint between data $x$ and latent variable $z$ in MAE $M=10$) is large, the models can achieve a competitive results of state-of-the-arts like pixelVAE  \cite{DBLP:journals/corr/OordKVEGK16} and Inverse Autoregressive Flow(IAF)  \cite{kingma2016improving}. When set the information constraint $F$ or $M$ to zero, we can observe that the results are comparable to the plain VAE, but less competitive than the former models.

\begin{table}[tbp]
\centering
\caption{Test negative log-likelihood (NLL) estimates for different models on MNIST}
\label{tab:nll}
\begin{tabular}{c|cccccccc}
\hline
             & VAE\citep{journals/corr/KingmaW13} & pixelVAE\citep{DBLP:journals/corr/OordKK16} &  IAF\citep{kingma2016improving} \\ \hline
NLL  &  85.56 & 79.21  &  79.85    \\ \hline
     & MAE ($M=10$) & MAE ($M=0$) & FAE ($F = 20$) & FAE ($F = 0$)  \\
\hline 
 NLL  &  80.86 &  81.58 & 79.30 & 83.24 \\
\hline
\end{tabular}
\end{table}

We put the former models in the FS information plane and draw the ``NJ curve'' for the Gaussian variable (where $\mathcal{N}(\mathbfcal{Z}|\mathbfcal{X}) \cdot tr(\mathbfcal{J}(\mathbfcal{Z}|\mathbfcal{X}))=K$) in the left subfigure of Figure~\ref{fig:FS}. According to the illustration, we can observe the trade-off between the Fisher information and entropy power in VAE. When the Fisher information elevates, the corresponding entropy power abases and vice versa. When the dependence between data and latent code is higher, where we set larger information constraint $F$ or $M$ in $q_{z|x}$, the corresponding models appear in the bottom-right corner in the FS plane. In the contrary, the models that contains less information in latent code appear in the upper-left corner, for instance, pixelVAE, which was reported to ignore the latent code \cite{chen2016variational} appears nearby FAE and MAE with $F = M = 0$. It is also interesting to notice that the inverse autoregressive flow VAE \cite{kingma2016improving} is beyond the curve. This is due to the IAF transforms the posterior into a more complex distribution from the simple Gaussian distribution. This phenomenon gives us the inspiration to improve the Fisher information and entropy power at the same time. That is to apply a more complex and a more proper distribution assumption for the modeling. 

From this plane, it is not hard to learn that we can vary the Fisher information constraint $F$ and ``move" on this curve. The upper-left corner indicates a less informative code, while the bottom-right indicates a more informative code.
In the right sub-figure of Figure \ref{fig:FS}, the training process of a FAE is also visualized in the FS plane. We plot the location of different epochs in FS plane. It is obvious that in FS plane the training process is intrinsically moving along the ``NJ curve" from upper-left side to the bottom-right side.  In fact, for most models, the goal is to move further in the bottom-right corner, thus we get better knowledge about data. Setting constraint of the Fisher information means to tell the model how much information we can transfer from data to the latent code, which can affect how far we can move to the right side along this curve.



\subsubsection{Effects of Fisher Information and Entropy Power}
The former part discusses the characteristics of different models and the corresponding performance. We are still curious about the effects of Fisher information and entropy power in encoding: when we should keep larger Fisher information and when we should keep larger entropy power? In this part, we show different effects when varying Fisher information and entropy power.

\begin{figure}
 \centering
 \includegraphics[width=\textwidth]{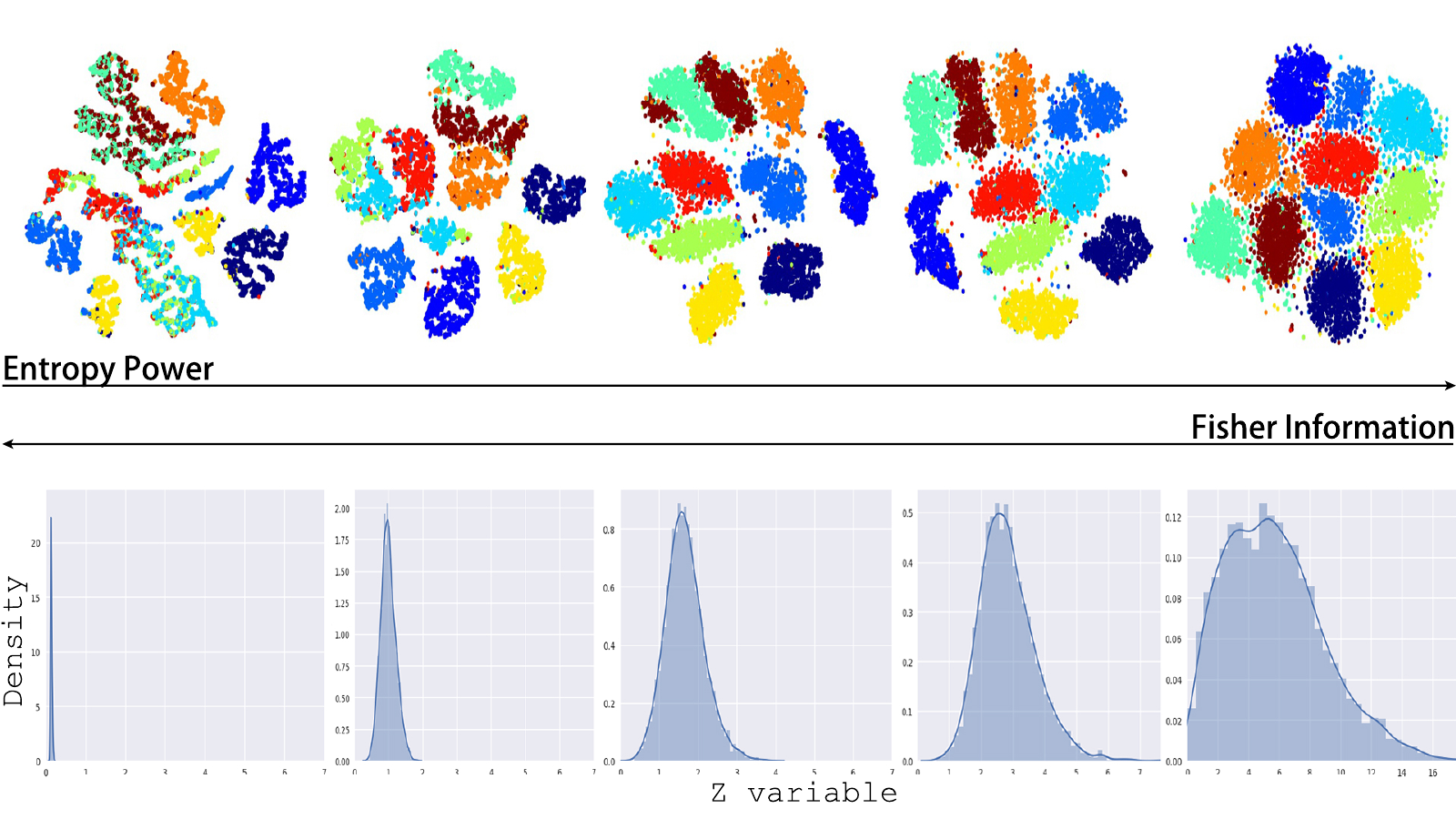}
\caption{\textbf{From left to right:} Visualization of latent code embedding (\textbf{upper}) and distribution of $q_{\phi}(z)$ (\textbf{lower}) as entropy power increases (FI decreases).}\label{fig:entro_fi}
\end{figure}

We set the latent code size to 10, and gradually increase the value of $F$ in FAE (from 0 to 20). The embedding of latent variable $z$ is visualized with T-SNE \cite{maaten2008visualizing}; the distribution of $q_{z|x}$ is visualized by sampling $z_i$ from $q_{z|x}$ and count the norm of normalized $z_i$, \textit{i.e.} $\Vert z_i - \overline{z}\Vert$. The results are presented in Figure \ref{fig:entro_fi}.

In Figure \ref{fig:entro_fi}, from left to right, the entropy power increases while the Fisher information decreases. As the entropy power increases, the latent variable embedding becomes more and more expanded in the latent space, where we can observe the clusters become more and more identifiable; while as the Fisher information increases, the embedding becomes more constrained to a smaller space. When observing the distribution $q_{z|x}$, it is obvious that the distribution is more centered when Fisher information is larger, while the distribution swells with larger variance when the Fisher information is abased.

As discussed, the Fisher information will control the variance of the encoding distribution. We can easily find out in Figure \ref{fig:entro_fi}, the variance of the distribution is getting smaller as FI increases. Intuitively, when VAEs encode the data, if we assign a large Fisher information constraint, the encoding variance is compressed to be smaller, thus the hashing cost is smaller and facilitates the model in distribution fitting. In the contrary, we can set a larger entropy power (or a smaller Fisher information) leaves more uncertainty to the encoding space, thus the latent code grabs the most common information from data points. This helps assemble data points in tasks like classification. 

In brief, we conclude the characteristics of large Fisher information and entropy power:
\begin{itemize}
    \item Large Fisher information provides with a more refined encoding mechanism, which ensures the latent code contains more detailed information. 
    \item Large entropy power provides with a more coarse encoding mechanism, which helps in global information extraction.
\end{itemize}
Larger Fisher information is thus helpful in learning of detailed features, high quality reconstruction, \textit{etc.}; while larger entropy power is helpful in classification, generalization, \textit{etc.}

\subsection{Qualitative Evaluation}
In this section, we present some qualitative results to provide an intuitive visualization. This will help us better understand the characteristics of Fisher information. 

 We present some reconstruction samples of FAE with large and small Fisher information constraint $F$ in Figure \ref{fig:rec}. As shown, the samples reconstructed with large $F$ provide with more pixel details and are more similar to the real images. This is especially more obvious in the case of SVHN, where we can observe more clear texture compared to the one reconstructed with larger constraint $F$. In the contrary, we can find some blurry samples reconstructed with small $F$ on MNIST. The blur is more obvious among reconstructed samples from SVHN.

\begin{figure}[tb]
 \centering
\begin{subfigure}{.32\textwidth}
\centering
         \includegraphics[width=\textwidth]{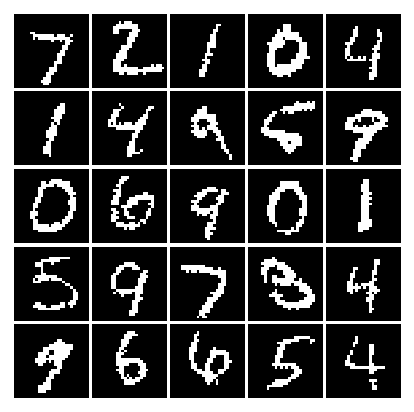}    
\end{subfigure}
\hfill
 \begin{subfigure}{.32\textwidth}
 \centering
         \includegraphics[width=\textwidth]{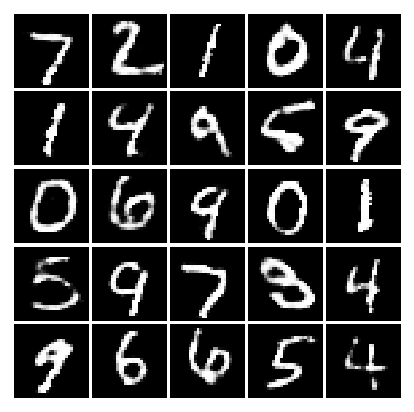}
\end{subfigure}
\hfill
 \begin{subfigure}{.32\textwidth}
 \centering
         \includegraphics[width=\textwidth]{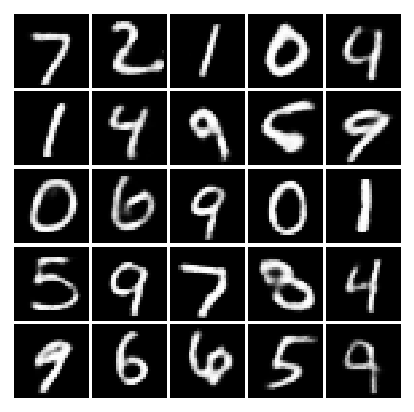}
\end{subfigure}
\begin{subfigure}{.32\textwidth}
\centering
         \includegraphics[width=\textwidth]{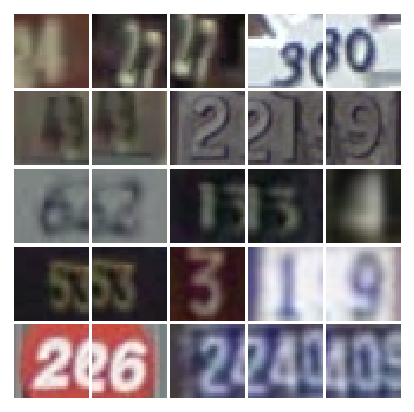}
         \subcaption{Real images}
\end{subfigure}
\hfill
 \begin{subfigure}{.32\textwidth}
 \centering
         \includegraphics[width=\textwidth]{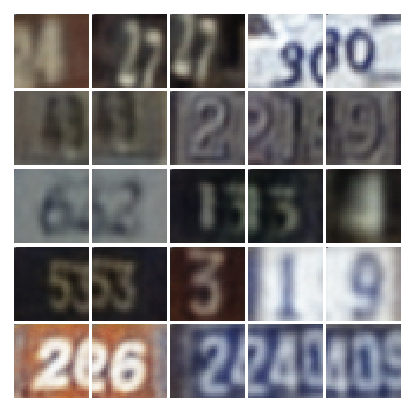}
         \subcaption{Reconstructions ($F$=20)}
\end{subfigure}
\hfill
 \begin{subfigure}{.32\textwidth}
 \centering
         \includegraphics[width=\textwidth]{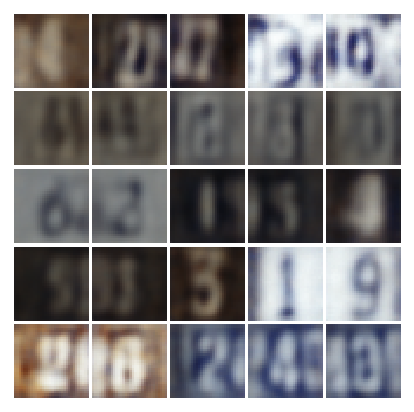}
         \subcaption{Reconstructions ($F$=0)}
\end{subfigure}
\caption{\textbf{From left to right}: \textbf{Real} images from test sets of MNIST (\textbf{upper}) and SVHN (\textbf{lower}, best viewed in color); images reconstructed by FAE with \textbf{large} ($F=20$) and \textbf{small} ($F=0$) Fisher information.}
\label{fig:rec}
\end{figure}

\begin{figure}
 \begin{subfigure}{\textwidth}
 \centering
         \includegraphics[width=\textwidth]{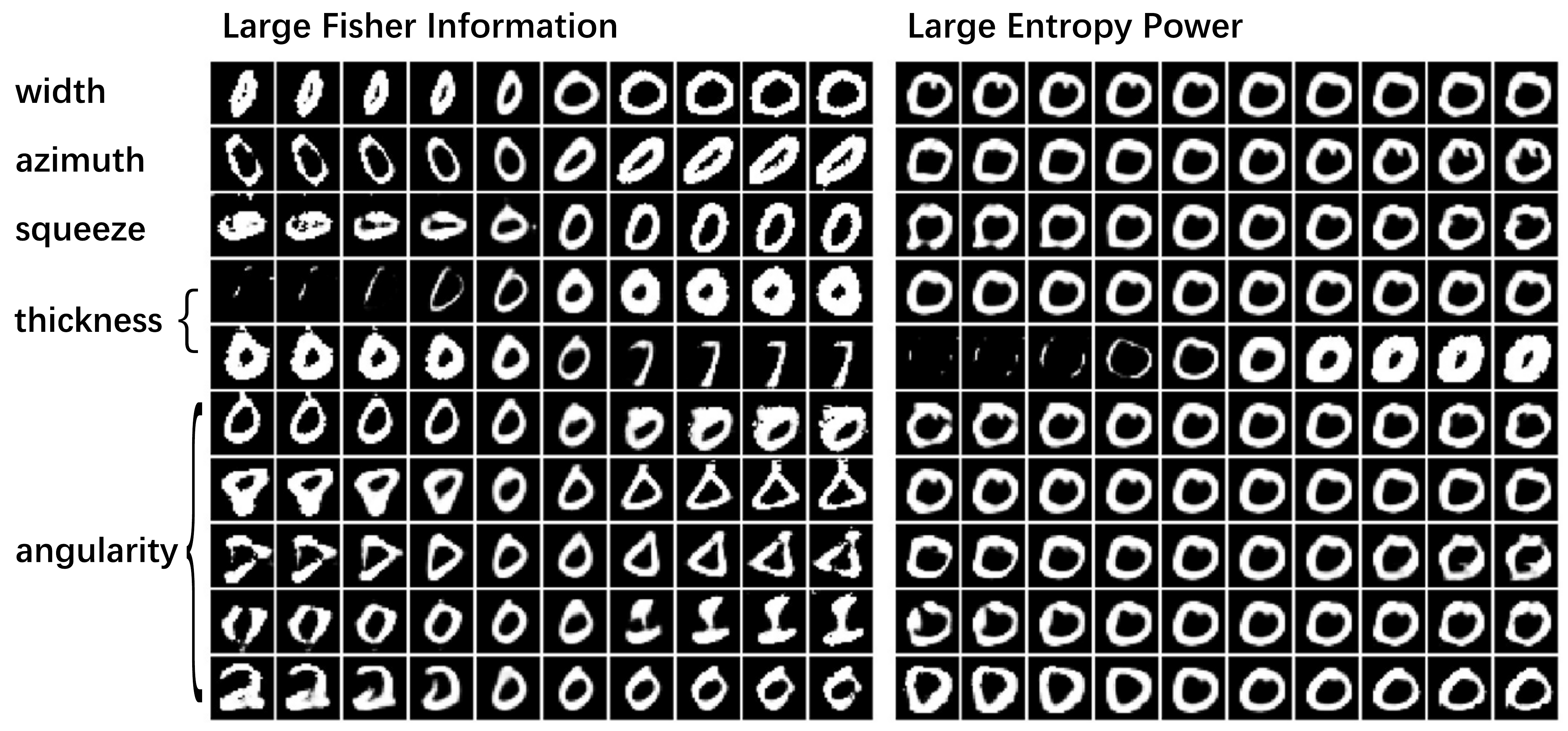}
\end{subfigure}
 \caption{FAE adjust Fisher information value to learn disentangle features in latent variable. (\textbf{left}: $F=20$; \textbf{right}: $F=0.1$)  }\label{fig:disentagle}
\end{figure}

We also present some qualitative results on disentangled feature learning in the latent variable in Figure \ref{fig:disentagle}. We respectively train two FAE model with $F = 20$ and $F = 0.1$; then we reconstruct samples by traversing over the latent variable and visualize the corresponding results. The traversal on 10 dimension of the latent variable $z$ is over the [-10, 10] range. As shown, the FAE with large $F$ learns a better disentangled representation of the generative factors.  The latent variable is encoded in a more refined mechanism, where we can distinguish the character that each dimension controls. For example, the first line presents the variance of the width of the digit zero; the second line represents the variance of the inclination direction of the digits, \textit{etc.} However, when we set $F = 0.1$, which indicates a large entropy power in the latent variable, the disentangled representation is not obvious as the former. This refers to the characteristic that we discussed in the previous sections: larger entropy power helps the model absorb similar attributes to make a high-level summary of data. 

\begin{table}[tbp]
\centering
\caption{Negative log-likelihood (NLL) estimates on reversed MNIST}
\label{tab:inorder nll}
\begin{tabular}{c|cccccccc}
\hline
      & FAE ($F = 20$) & FAE ($F = 10$) & FAE ($F=5$)  & FAE ($F=0.1$) \\
\hline 
Train NLL & 94.48  & 95.12 & 95.89  & 96.67 \\
\hline 
Test NLL & 164.56  & 134.94 &  109.19  & 108.37  \\
\hline
\end{tabular}
\end{table}
\subsection{Generalization of using FAE}
In this section, we conduct a series of experiments on the reversed MNIST dataset (which mostly consists of data with label ``0", as described in section \ref{sec:ex setting}) to describe a scenario, where we should constrain the Fisher information to obtain larger entropy power.

We trained FAE on this dataset with different $F$. The results of test log-likelihood are presented in Table \ref{tab:inorder nll}. From the table, the train NLL is shown to decrease as $F$ increases, while the test NLL increases. This phenomenon indicates the model suffers from over-fitting on the dataset.

\begin{figure}
 \begin{subfigure}{\textwidth}
 \centering
         \includegraphics[width=\textwidth]{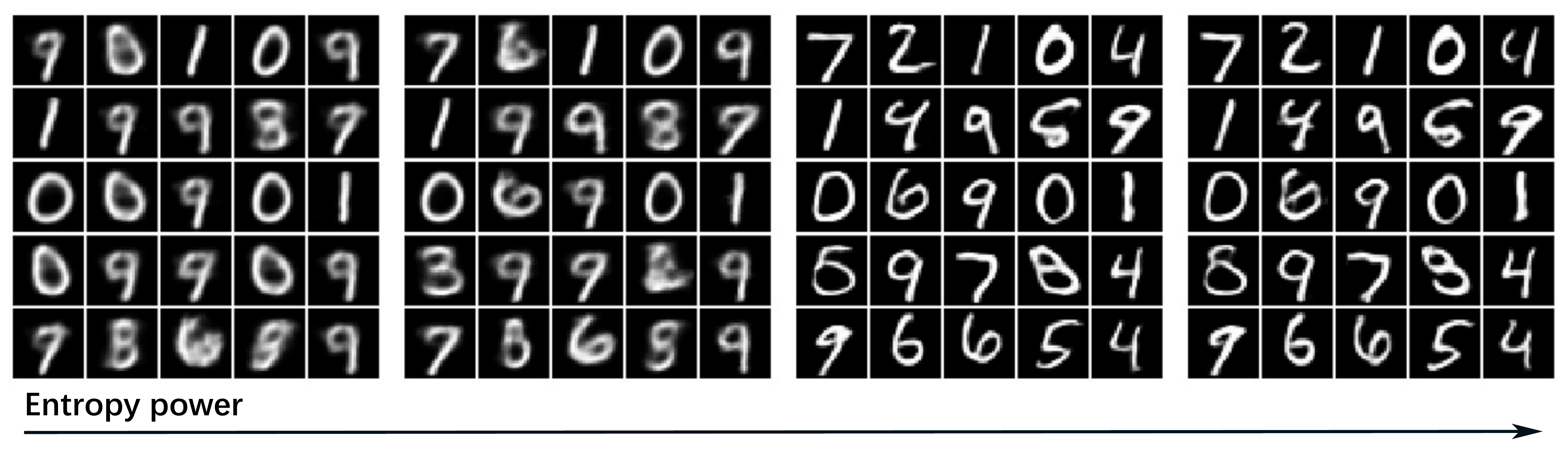}
\end{subfigure}
 \caption{The reconstructed samples by FAE with increasing entropy power. }\label{fig:inorder}
\end{figure}

In Figure \ref{fig:inorder}, the reconstructed samples are presented. When Fisher information is large, the model tends to well fit data. However, since the most part of training data are digit zero, the model mainly captures the attributes of digit ``0". In test cases, the model will reconstruct samples in a way that contains some attributes of zero. As we constrain the Fisher information, the model has better capacity of generalization with larger entropy power to overcome the over-fitting.

\subsection{Discussion: How to Benefit FS Information}
From the previous parts, we can learn that larger Fisher information helps in precise distribution modeling, disentangled feature learning, and achieve a good reconstruction accuracy; while larger entropy power leaves more uncertainty in the model for generalization and enables to handle the over-fitting. 

In real world, when we train a model on a dataset, it is important to estimate the uncertainty of the given dataset; for example, whether the data distribution is skewed, \textit{etc.}. When the dataset is believed to be complete, or to be representative for the future incoming data, larger Fisher information can be preferred in the model to refine the learning. Otherwise, we should warn the model with uncertainty in the future, so it can
get a better generalization capacity to handle the ``surprisal" in the future.

\section{Conclusion}
Based on the uncertainty property between Fisher information and Shannon information, in this paper, we apply the Fisher-Shannon plane to study VAEs in a joint view of these two quantities. In our study of VAEs in Fisher-Shannon plane, these information quantities are demonstrated related to the representation learning and likelihood maximization; the trade-off between Fisher information and Shannon information is shown to result in different characteristics of VAEs. We further propose the Fisher Auto-Encoder for the information control by different Fisher information constraints. In our experiments, we demonstrate the complementary characteristics of Fisher information and Shannon information and provide with a novel understanding of VAEs; we also justify the effectiveness of FAE in information control, high-accuracy reconstruction and non-informative latent code resistance.

\bibliographystyle{plainnat}
\bibliography{nips_2018.bib}

\newpage

\appendix
\section{Non-parametric Fisher information}
The Fisher information we formulated in section \ref{sec:ineq} is a non-parametric version. In the original definition of Fisher information, it is formulated as \citep{STAM1959101}:
\begin{align} \label{eq:parameter FI}
\begin{split}
\mathcal{J}_{\theta}(X) & =  - \int_{x}p_\theta(x)\frac{\partial^2}{\partial \theta^2}p_\theta(x) dx \\
                        & =  \int_{x}\frac{1}{p_\theta(x)}(\frac{\partial}{\partial \theta}p_\theta(x))^2 dx
\end{split}
\end{align}
where $x$ can be multi-dimensional. In \citep{STAM1959101}, if $p_\theta(x)$ only depends on $x - \theta$, for example, the gaussian distribution \textit{w.r.t.} mean $\mu$, then $\theta$ can be dropped from $J_{\theta}(X)$ to become a non-parametric version.
\begin{align} \label{eq:non-parameter FI}
\begin{split}
\mathcal{J}(X) = \int_{x}\frac{1}{p(x-\theta)}(p^\prime(x-\theta))^2 dx
\end{split}
\end{align}

In reality, we can regard the distribution parameter as a variable and formulate the parametric distribution as a posterior:
\begin{equation}
    p_\theta(x) = p(x|\theta)
\end{equation}
Then we can manipulate the prior of $\theta$ to find a distribution $q$ that satisfies:
\begin{equation}
    p(x,\theta) = p(\theta)p(x|\theta) = q(x-\theta)
\end{equation}
Finally, we transform a distribution from parametric version into non-parametric version. This transformation needs to satisfy three conditions according to \citep{STAM1959101}:
\begin{enumerate}
    \item $q$ is strictly positive
    \item $q$ is differentiable
    \item The integral \eqref{eq:non-parameter FI} exists
\end{enumerate}

\section{Minor Characteristic: Stability in Learning}
In terms of optimization for VAEs, the stability of parameter estimation is also a concern~\citep{kingma2016improving,zhao2017infovae}, which also reflects a minor characteristic of using Fisher information. In~\citep{deg2018zheng}, the instability of parameter will lead to degeneration when VAE's architecture is deep and is linked to the Fisher information loss. Here, we present the resistance of degeneration with the advantage of Fisher information regularizer in VAEs.

\begin{figure}
 \begin{subfigure}{\textwidth}
 \centering
         \includegraphics[width=\textwidth]{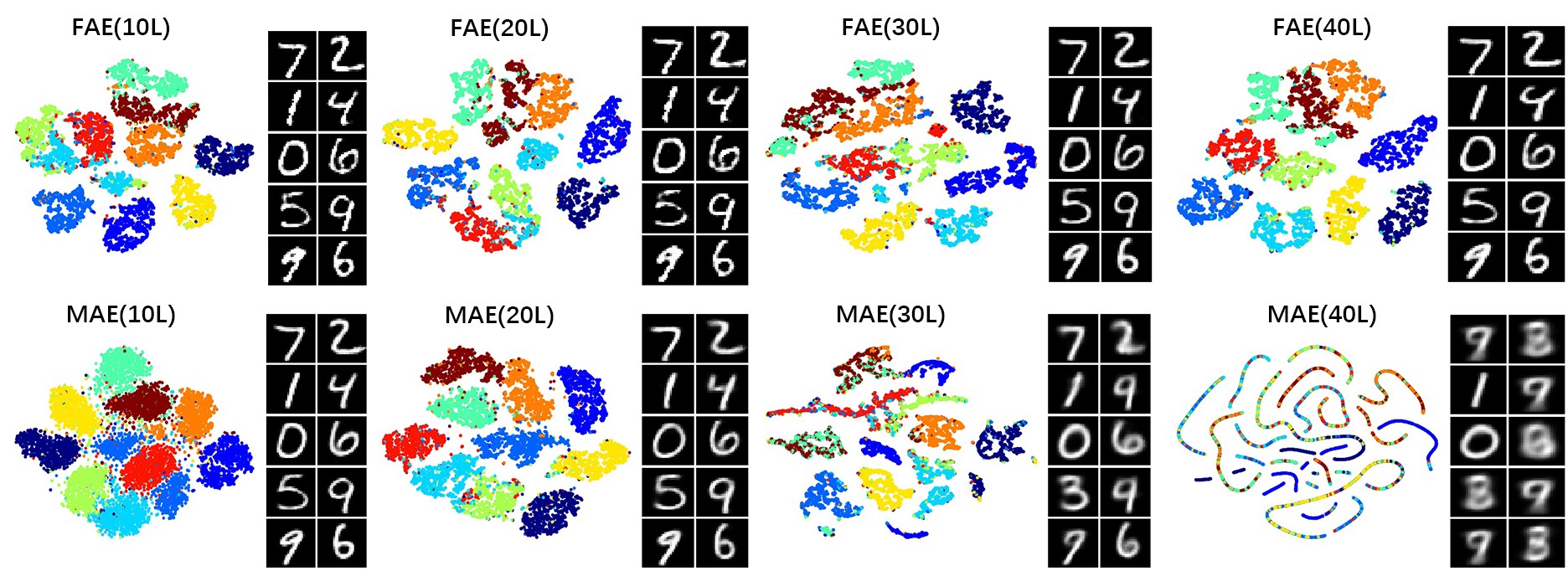}
\end{subfigure}
 \caption{FAE and MAE analysis in different model complexity (best view in color). Visualization of learned latent space and sample reconstructed by FAEs  (\textbf{upper}) and MAEs.  (\textbf{lower}) }\label{fig:deg}
\end{figure}

Specifically, we extend the encoder and the decoder to 20, 30, 40 layers (marked as 20L, 30L, 40L) respectively to observe the performance of FAE and MAE in the context of degeneration. The performance is illustrated in Figure \ref{fig:deg}. As can be seen, the latent space and the reconstruction is similar to those in normal situation with the network going deep. Fisher information captures the variability of the gradient and can alleviate the degeneration in very deep networks. Instead, MAE, although it can also control information in encoding mechanism, it is risky in gradient vanishing. It thus causes inaccurate parameter estimation and results in degeneration in the latent space and reconstruction.

\section{Supplementary experiment results on SVHN}

\begin{figure}[hbp]
 \centering
\begin{subfigure}{.3\textwidth}
\centering
         \includegraphics[width=\textwidth]{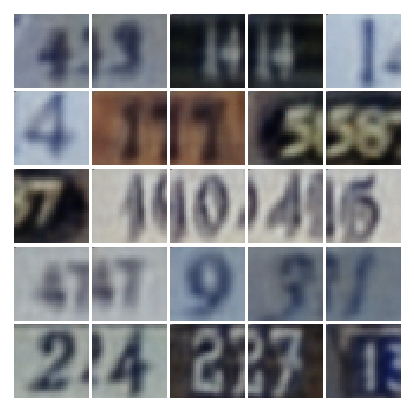}    
\end{subfigure}
\hfill
 \begin{subfigure}{.3\textwidth}
 \centering
         \includegraphics[width=\textwidth]{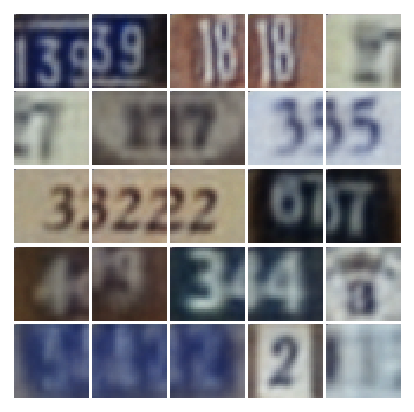}
\end{subfigure}
\hfill
 \begin{subfigure}{.3\textwidth}
 \centering
         \includegraphics[width=\textwidth]{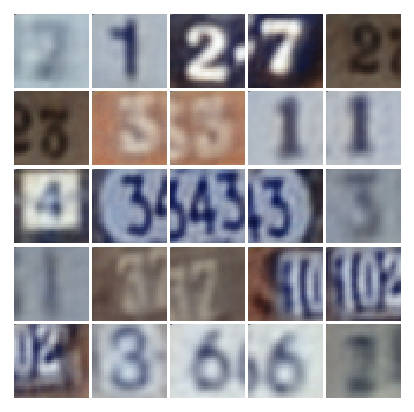}
\end{subfigure}
\begin{subfigure}{.3\textwidth}
\centering
         \includegraphics[width=\textwidth]{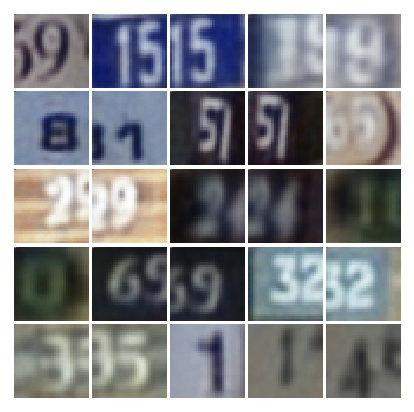}    
\end{subfigure}
\hfill
 \begin{subfigure}{.3\textwidth}
 \centering
         \includegraphics[width=\textwidth]{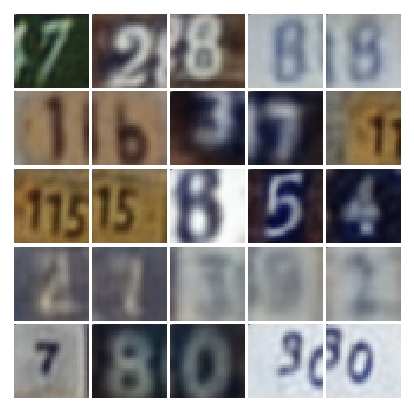}
\end{subfigure}
\hfill
 \begin{subfigure}{.3\textwidth}
 \centering
         \includegraphics[width=\textwidth]{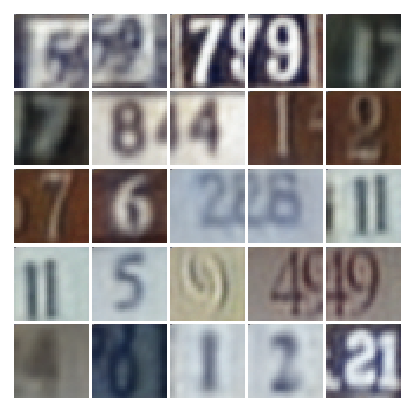}
\end{subfigure}
\caption{Samples reconstructed by FAE.}
\label{fig:rec}
\end{figure}

\end{document}